\documentclass[10pt,conference]{IEEEtran}
\usepackage{cite}
\usepackage{algorithmic}
\usepackage{array}
\usepackage{subfigure}
\usepackage[cmex10]{amsmath}
\usepackage{amsmath}
\usepackage{physics}
\ifCLASSINFOpdf
	 \usepackage[pdftex]{graphicx}
\else

\fi
\usepackage{cite}

\usepackage{fancyhdr}
\begin{document}

\title{On the Modeling of Error Functions as High Dimensional Landscapes for Weight Initialization in Learning Networks}
\author{\IEEEauthorblockN{Julius$^1$, Gopinath Mahale$^2$, Sumana T.$^2$, C. S. Adityakrishna$^3$}
\IEEEauthorblockA{\small \{$^1$Department of Physics, $^3$Department of Computer Science\}, BITS-Pilani Hyderabad, India, $^2$CAD Lab, Indian Institute of Science, India \\
Email: \emph{ \{juliuswombat,aditya.chivukula\}@gmail.com  \{mahalegv, sumana.t\}@cadl.iisc.ernet.in}}
}

\maketitle

\begin{abstract}
Next generation deep neural networks for classification hosted on embedded platforms will rely on fast, efficient, and accurate learning algorithms. Initialization  of weights in learning networks 
has a great impact on the classification accuracy. In this paper we focus on deriving good initial weights by modeling the error function of a deep neural network as a  high-dimensional landscape. We observe that due to the inherent  complexity in its algebraic structure, such an error function may  conform to general results of the statistics of large systems. To this  end we apply some results from Random Matrix Theory to analyse these  functions. We model the error function in terms of a Hamiltonian in  N-dimensions and derive some theoretical results about its general behavior. These results are further used to make better initial guesses of weights for the learning algorithm. 
\end{abstract}

\section{Introduction}
Machine learning is a discipline in which correlations drawn from  samples are used in adaptive algorithms to  extract critical and relevant information that helps in classification. The interplay between the formulations of learning models in the primal/dual spaces has a great impact in the theoretical analysis and in the practical implementation of the system, more so when emerging embedded platforms play host to a variety of classification systems and applications. Clearly, the demand on performance efficiency and accuracy of such machine learning system is of paramount interest.

Learning can be supervised, unsupervised or even a combination of the two. In face recognition systems for instance, machine learning is typically supervised since it is trained using a sample set of faces. Whereas, in big data analytics, machine learning is unsupervised since there is no {\it a priori} knowledge of the features/information associated with the data. In the terminology of machine learning classification is considered an instance of supervised learning, i.e. learning where a training set of correctly identified observations is available (i.e. the training set is labelled). In unsupervised learning, a model is prepared by deducing structures present in the input data (which is either not labelled or no {\it a priori} labelling is known) and project them as general rules. This could mean identifying a mathematical method/process for data organization that systematically reduces redundancy. The corresponding unsupervised procedure is known as clustering, and involves grouping data into categories based on some measure of inherent similarity or distance.  In semi-supervised learning,
input data is a mixture of labelled and unlabelled samples. There is a desired prediction problem but the model must learn the structures to organize the data as well as make predictions.

\begin{figure}[!b]
\centering
\subfigure[A single neuron\label{fig: neuron}]{\includegraphics[width=3.0cm, height=2.5cm]{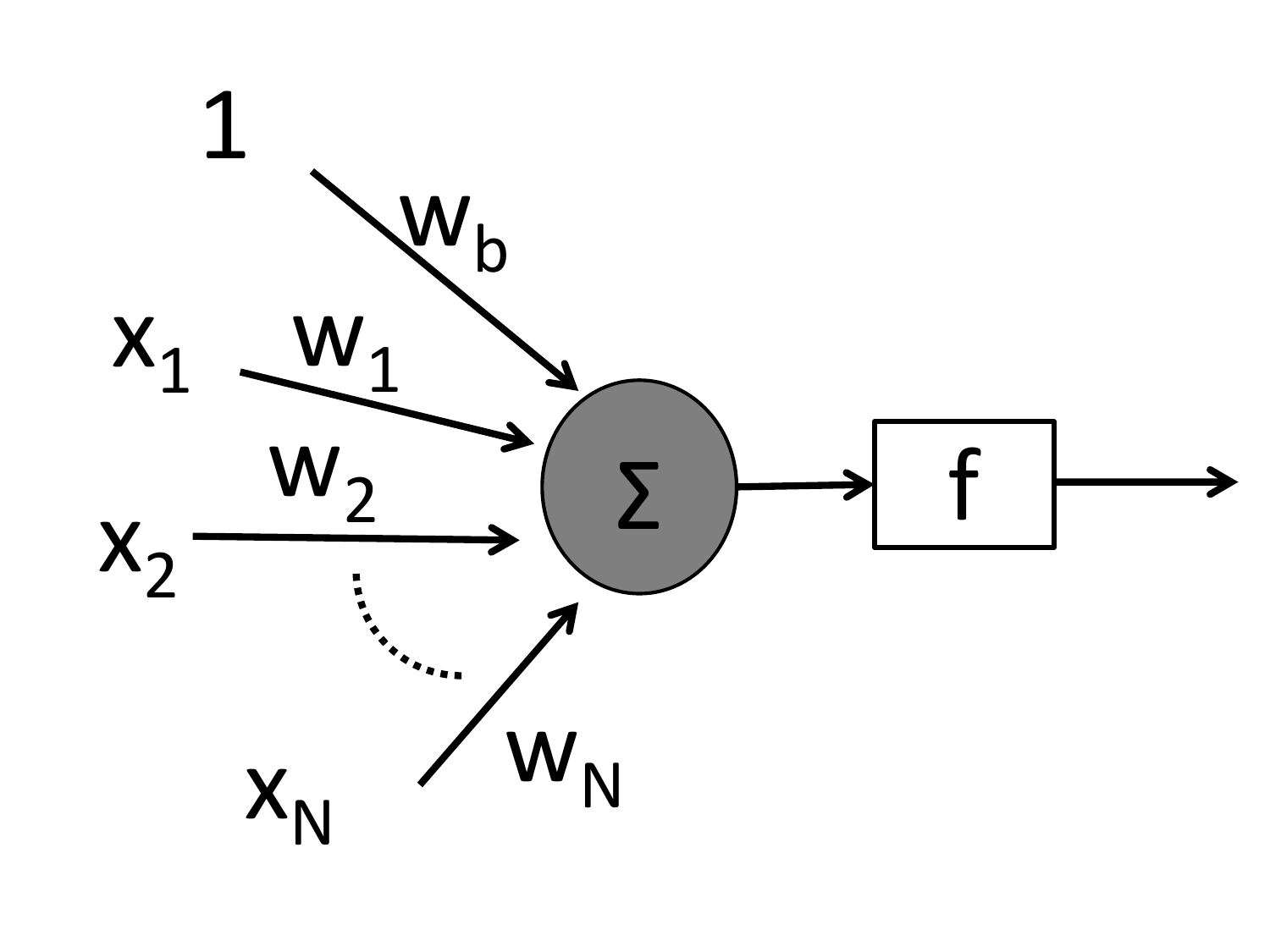}}
\subfigure[A three layer feed forward neural network\label{fig: ffnn}]{\includegraphics[width=5.5cm, height=4.0cm]{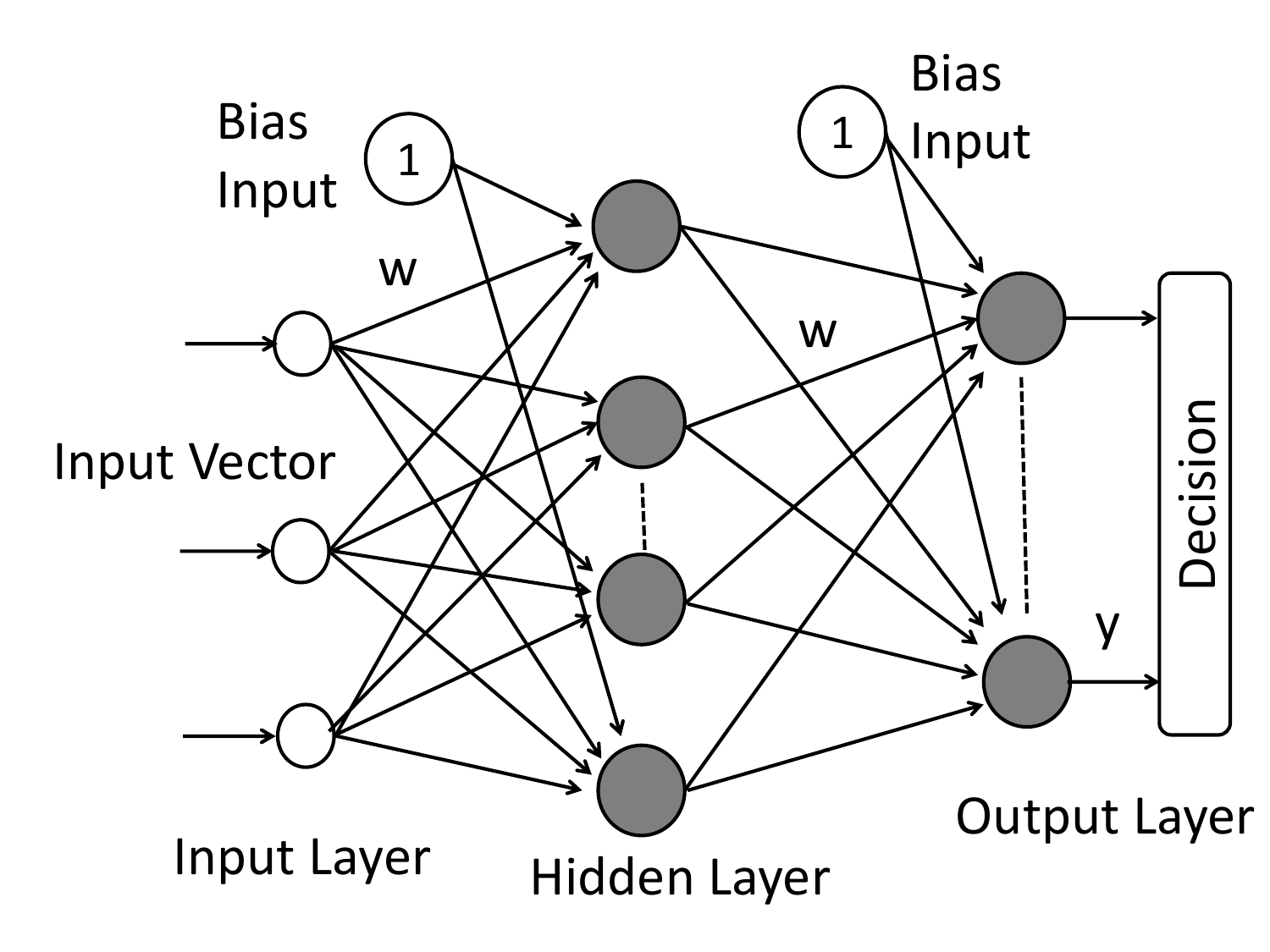}}

\caption{A single neuron and a feed forward neural network}
\label{fig:NN}
\end{figure}

Neural networks (NNs) are one of the major developments in the field of machine learning. The popularity of NNs is due to its substantial learning capacity and adaptability to various application domains. The building blocks of a NN are called neurons that act as processing nodes. Such nodes arranged in layers make the network. The layers are called input, hidden or output layer based on their function and visibility to the programmer. These layers are interconnected by synaptic links that have  associated synaptic weights. A pictorial representation of a neuron and a feed forward neural network is shown in Fig. \ref{fig: neuron} and Fig. \ref{fig: ffnn} respectively. Training a NN refers to tuning the synaptic weights to implement a given function. The function computed by each neuron is 
\begin{equation} \label{eqn: output_node}
y = f \left(w_b + \sum_{i = 1}^{N}w_{i}\times x_{i}\right)
\end{equation}
where, $N$ is the dimension of the input sample. $x_i$ and $w_{i}$ are $i^{th}$ element of the input sample and weight vector respectively. $w_b$ is weight associated with the bias input as shown in Fig. \ref{fig: neuron}. $f$ is a differentiable non-linear function. Some of the popular non-linear functions used are $sigmoid$, $tanh$, $ReLU$ etc.  During training of this neuron, samples $x_{tr}$ from the training database $X_{tr}$, each associated with labels $y_{tr}$ are used to train the synaptic weights. This tuning of weights is performed to minimize the error function which is a function of difference between the predicted output and the actual output given by
\begin{equation}
E = \frac{1}{N_{samp}} \times \sum_{i = 1}^{N_{samp}} \left( y_{tr_i} - f \left(w_b + \sum_{j = 1}^{N}w_{j}\times x_{tr_{i,j}}\right)\right)^2
\end{equation}
 where $x_{tr_{i,j}}$ is the $j^{th}$  element  of $i^{th}$ training sample and $N_{samp}$ is the number of training samples. The error function is a high dimensional landscape, which needs to be explored for its minima. 
 
A single neuron can be trained using standard procedures such as gradient descent that updates the weights based on gradient of the error function. When it comes to the training of a multi-layer feed forward NN, the gradient descent involves layer-wise computation of gradients and tuning the weights accordingly. This method is popularly known as the Back propagation. To train a multi-layer NN using back propagation, there are two main design parameters to be chosen. First, the learning rate  which can be visualized as a step size in the search for minima in the error landscape. During training, dynamic update of learning rate has shown to perform better learning in practical examples. The second, but more important, design parameter to be chosen is the initial synaptic weights to start the back propagation. The initialized weights have shown to affect the number of iterations required for the convergence along with the classification performance of the trained network \cite{init}. Weight initialization methods such Xavier method \cite{xavier} and Nguyen-Widrow method \cite{NW} are being used in deep learning frameworks like Caffe \cite{caffe} and Matlab Neural Network Toolbox\cite{matlab} for faster and efficient learning. Majority of these deep learning frameworks employ  stochastic gradient descent algorithms to arrive at the optimal weight vector for accurate classification.

Deep Neural Networks (DNNs) \cite{deep} have recently emerged as the area of interest for researchers in the field of machine learning. The strength of DNN lies in the multiple layers of neurons that together are capable of representing a large set of complex functions. Although we see numerous applications of DNN, training DNNs has always been a challenge due to the large number of layers. In addition, these very deep networks have witnessed the problem of vanishing gradients \cite{vanishing}, that has encouraged the researchers to explore better methods of initializations. A good weight initialization has shown to play a crucial role in achieving better minima with faster learning in DNNs. Therefore, there is a need for better weight initialization methods that will play a major role in training emerging very deep neural networks.

In this paper we explore statistical methods from Random Matrix Theory (RMT)\cite{Edelman} for large systems, and apply these concepts to explore High Dimensional Landscapes of Error Functions for fast learning. While such methods have been applied to problems in Physics to study complex energy levels of heavy nuclei, financial analytics for stock correlations, communication theory of wireless systems, array signal processing, this is for the first time (to the best of our knowledge) that  such a method is being applied to learning systems. The rest of the paper is organized as follows. In section \ref{sec: RMT} we give a brief introduction to the RMT and its applicability in learning systems in the context of prior  work. We refer to analysis of RMT and results in the literature which are the basis for development of our approach for faster learning. In section \ref{sec: theory}, we describe our approach that applies RMT to the problem of learning in neural networks. Section \ref{sec: analysis} contains analysis of different parameters in RMT to improve the learning ability of the network along with related results to support our theory. We conclude in section \ref{sec: conclusion}.

\section{Some History: Random Matrix Theory and Related Work} \label{sec: RMT}
 \label{sec:wig}
  Lately Random Matrix Theory (RMT) has been applied effectively in various fields of science and engineering \cite{Edelman}. The fact that little knowledge of RMT is sufficient for its application \cite{Edelman} has encouraged considerable research work towards exploring applicability of RMT in different application domains.

We present two fundamental results of RMT that appear again and again in many of the models characterized by random matrices.

\subsection{The Semicircle Law}
\label{sec:semi}
Wigner's Semicircle Law \cite{wigner} can be stated as follows
\textit{Consider an ensemble of $N \cross N$ real symmetric matrices with independent identically distributed random variables from a fixed probability distribution $p(x)$ with mean 0, variance 1, and other moments finite. Then as $N \rightarrow \infty$}
\begin{align}
\label{eqn:wig1}
\mu_{A,N}(x) = 
\begin{cases}
\frac{2}{\pi}\sqrt{1 - x^{2}} & \text{if } \abs{x}\leq 1 \\
0 & \text{otherwise}
\end{cases}
\end{align}
\textit{In other words, the sum of normalized eigenvalues in an interval $[a,b] \subset [-1,1]$ is found by integrating the semicircle over that interval.}

The semicircle law provides the requisite connection between the eigenvalues of a random matrix and the moments of an ensemble of random matrices. 

\subsection{The Tracy Widom Law}
\label{sec:TWLaw}

Problems regarding the motion of a particle in a high-dimensional landscape occur throughout physics in various disciplines such as Spin-glass theory, String theory, the theory of Supercooled liquids, etc.~\cite{majumdar}. The dynamics of a system in an $N$-dimensional potential can be described by 
\begin{equation}
	\label{eqn:TW}
	\frac{dy_{i}}{dx} = -\nabla_{i}V
\end{equation}
where $V = V(x_{1}, x_{2}, ..., x_{n})$ is the functional form of the potential of interest. A high-dimensional landscape is characterized by its stationary points. Stationary points are points on the landscape where a particle moving on it is at equilibrium. At stationary points, the gradient of the potential vanishes. The nature of the stationary points is determined by the laplacian of the potential which is given by the eigenvalues of the Hessian matrix of the system. A matrix element of the Hessian matrix is defined as
\begin{equation}
	\label{eqn:TW2}
	H_{ij} = \nabla^{2}_{ij}V
\end{equation}
The probability of finding a local minima is given by $P(\lambda_{1} < 0, \lambda_{2} < 0, ..., \lambda_{n} < 0)$.
This is equivalent to finding the probability that the maximal eigenvalue $\lambda_{max} < 0$. The study of the maximal eigenvalue of a random matrix is thus of appreciable interest in disciplines that study high-dimensional landscapes.
Tracy and Widom \cite{tracy} showed in 1994 that the distribution of the maximal eigenvalue of an ensemble of random matrices is given by
\begin{equation}
	\label{eqn:TW4}
	P(\lambda_{max} \leq w, N) = \mathcal{F}_{\beta}(\sqrt{2}N^{-2/3}(w - \sqrt{2}))
\end{equation}
where $\mathcal{F}_{\beta}$ is obtained from the solution to the Painlev\'{e} II equation. $\beta = 1$ corresponds to the Gaussian Orthogonal Ensemble (G.O.E.). The G.O.E. will be our ensemble of interest in this paper. Setting $\beta = 1$ the expression for $\mathcal{F}_{1}$(x) is given by
\begin{equation}
	\label{eqn:TW5}
	\mathcal{F}_{1}(x) = \exp(-\frac{1}{2}\int_{x}^{\infty}((y - x)q^{2}(y) + q(y))dy)
\end{equation}
where $q$ is given by
\begin{equation}
	\label{eqn:TW6}
	\frac{d^{2}y}{dx^{2}} = 2q^{3}(y) + yq(y)
\end{equation}
Here $N$ is the dimension of the random matrix under consideration. The Tracy Widom Law thus provides us with a powerful analytical tool in the study of minima of a large random landscape: the main subject of this paper.
\subsection{High Dimensional Landscapes}
\label{sec:HDLand}

 Fyodorov et. al. \cite{fyod1} state that finding total number of stationary points in a spatial domain of random landscape is difficult and no efficient techniques are available to perform the task. However, it is possible to perform such calculation for Gaussian fields $\mathcal{H}(x)$ which have isotropic covariance structure, i.e., covariance only dependent on the Euclidean distances $|x_1 - x_2|$. The estimation of stationary points in a spatial domain is equivalent to evaluating the mean density of eigenvalues of the Gaussian Orthogonal Ensemble (G.O.E.) of real $N \times N$ random matrices which has a well established closed form expression \cite{mehta}.

We refer to concepts in \cite{fyod1} to build a base for our theory elaborated in section \ref{sec: theory}. We consider a random Gaussian landscape
\begin{equation}
\label{eqn:ham1}
\mathcal{H} = \frac{\mu}{2}\sum_{i = 1}^{N}n_{i}^{2} + V(n_{1}, n_{2}, ..., n_{N})
\end{equation} 
where $\mu$ is a tuning parameter and $V$ is a random mean-zero Gaussian-distributed field with covariance given by 
\begin{equation}
\label{eqn:ham2}
\expval{V({\bf{m}}), V(\bf{n})} = Nf\left( \frac{1}{2N}\abs{\bf{m} - \bf{n}}^{2}\right)
\end{equation}
where $f(x)$ is a smooth function decaying at infinity.  Comparing $\mathcal{H}(x)$ with the Gaussian Orthogonal Ensemble (G.O.E.) and applying the tools \cite{fyod1} of RMT, one can count the average number of minima $\expval{\mathcal{N}_{m}}$ of $\mathcal{H}$. As discussed in \cite{fyod1}, $\expval{\mathcal{N}_{m}}$ is given by
\begin{equation}
\label{eqn:fyod1}
\expval{{\mathcal{N}}_{m}} = \left(\frac{\mu_c}{\mu}\right)^{N}\frac{2^{(N+3)/2}\Gamma(\frac{N+3}{2})}{\sqrt{\pi}(N+1)N^{N/2}}I_{N}\left(\frac{\mu}{\mu_{c}}\right)
\end{equation}
where $I_{N}(\mu/\mu_{c})$ is given by
\begin{equation}
\begin{split}
\label{eqn:fyod2}
I_{N}\left(\frac{\mu}{\mu_{c}}\right) = \int_{-\infty}^{\infty}e^{\left(\frac{s^{2}}{2}-\frac{N}{2}\left(s\sqrt{\frac{2}{N}} - \frac{\mu}{\mu_{c}}\right)^{2}\right) }
\\
\frac{d}{ds}\left({P}_{N+1}(\lambda_{max}\leq s)\right)ds
\end{split}
\end{equation}

Here $\mu_{c} = \sqrt{f^{\prime\prime}(0)}$. ${P}_{N}(\lambda_{max}\leq s)$ is the probability that the maximal eigenvalue of a standardized G.O.E. matrix $M$ is smaller than $s$. The Tracy-Widom Law \cite{tracy} gives us the following formula as $N\rightarrow\infty$.
\begin{equation}
\label{eqn:fyod3}
{P}_{N}\left(\frac{\lambda_{max}-\sqrt{2}N}{N^{-1/6}/\sqrt{2}}\leq s\right) \sim \mathcal{F}_{1}(s)
\end{equation}
where $\mathcal{F}_{1}$ is a special solution of the Painlev\'{e} II equation.
Substituting back into equation \ref{eqn:fyod2} we get
\begin{equation}
I_{N}\left(\frac{\mu}{\mu_{c}}\right) = \int_{-\infty}^{\infty}e^{h_{N}(t)}dt
\label{eqn:fyod4}
\end{equation}
where $h_{N}(t)$ is 
\begin{equation}
\label{eqn:fyod5}
h_{N}(t) = \frac{s_t^{2}}{2}-\frac{N}{2}\left(s_t\sqrt{\frac{2}{N}}-\frac{\mu}{\mu_{c}}\right)^{2} + \ln\mathcal{F}_{1}^{'}(t)
\end{equation}
where $s_t = \sqrt{2(N+1)}+t\frac{(N+1)^{-1/6}}{\sqrt{2}}$.
Thus as originally shown in \cite{fyod1} expressions \ref{eqn:fyod1}, \ref{eqn:fyod4} and \ref{eqn:fyod5} help to arrive at the average number of minima for the energy landscape defined by $\mathcal{H}$.

The explicit formula for ${P}_{N}(\lambda_{max}\leq s)$ is 
\begin{equation}
\label{eqn:fyod6}
{P}_{N}(\lambda_{max}\leq s) = \frac{Z_{N}(s)}{Z_{N}(\infty)}
\end{equation}
where $Z_{N}$ is
\begin{equation}
\begin{split}
\label{eqn:fyod7}
Z_{N}(s) = \int_{-\infty}^{s}d\lambda_{1}\int_{-\infty}^{s}d\lambda_{2} \cdots \int_{-\infty}^{s}d\lambda_{N} \\ \prod_{i<j}\abs{\lambda_{i} - \lambda_{j}}\exp(-\lambda_{i}^{2}/2)
\end{split}
\end{equation}
as given in \cite{fyod1}. Here $\lambda_i$ is the $i^{th}$ eigenvalue of $\mathcal{H}$. In equation \ref{eqn:fyod2} the limits of the integration are over the entire real line. From equations \ref{eqn:fyod6} and \ref{eqn:fyod7} we see that this manifests as an integral over all the possible minima for each $x_{i}$. Consider having finite limits to the integral in equation \ref{eqn:fyod2}. We see again from equations \ref{eqn:fyod6} and \ref{eqn:fyod7} that this would mean that we're searching for minima in the range defined by our new limits. We have thus defined an $N$-dimensional hypercube inside which we choose to search for minima. Introducing finite limits in equation \ref{eqn:fyod4} we get
\begin{equation}
\label{eqn:dens1}
I_{N}\left(\frac{\mu}{\mu_{c}}\right)\bigg{|}_{a}^{b} = \int_{a}^{b}e^{h_{N}(t)}dt
\end{equation}
with $h_{N}$ as defined in equation \ref{eqn:fyod5}. This gives
\begin{equation}
\label{eqn:dens2}
\expval{{\mathcal{N}}_{m}}_{a}^{b} = \frac{\mu}{\mu_{c}}^{N}\frac{2^{(N+3)/2}\Gamma(\frac{N+3}{2})}{\sqrt{\pi}(N+1)N^{N/2}}I_{N}\left(\frac{\mu}{\mu_{c}}\right)\bigg{|}_{a}^{b}
\end{equation}
Equation \ref{eqn:dens2} provides us a formula by which we can calculate the mean number of energy minima of the Hamiltonian $\mathcal{H}$ within a hypercube defined by the limits of the integral in the R.H.S.

\subsection{Related Work}
\label{sec:relatedwork}

In~\cite{init}, a Layer-sequential unit-variance (LSUV) initialization method for weight initialization, in connection with standard stochastic gradient descent (SGD) is proposed that leads to state-of-the-art thin and very deep neural nets. Through experimental validation the authors establish that the proposed initialization leads to learning of very deep nets that produces networks with test accuracy better or equal to standard methods and is comparable to complex schemes such as FitNets~\cite{fitnet} and Highway~\cite{deep_net}.

Authors in~\cite{greedy} propose greedy layer-wise unsupervised training strategy for deep multi-layer neural networks that targets initializing weights in a region near a good local minimum, giving rise to internal distributed representations that are high-level abstractions of the input, resulting in better generalization.

Authors in~\cite{Lecun} draw a parallel between machine learning problems, such as deep networks, and mean field spherical spin glass model in terms of finding the ground state energy of the mean field system's Hamiltonian ($\mathcal{H}$), though in reality the two systems are not mathematically analogous. They hint at the two systems being special cases of a more general phenomenon which is not yet discovered. Through experiments on teacher-student networks with the MNIST dataset, the authors report that both gradient descent and their proposed stochastic gradient descent methods (based on RMT) are equally efficient  in identical number of steps. 

As discussed so far,  RMT provides statistics about the eigenvalue spectrum of an arbitrary random matrix. Quantum theory treats all physical observables as operators. The energy of any physical system corresponds to the result of the action of the Hamiltonian operator $\mathcal{H}$ on the system. Quantum theory provides the formalism by which the Hamiltonian of a system can be realized as a Hermitian matrix. Thus finding the energy of a system boils down to solving the eigenvalue problem of the Hamiltonian matrix. Since the Hamiltonian is a large complicated infinite dimensional matrix it only makes sense to talk statistically. RMT provides us the necessary mathematical tools to do so.

In this paper, we explore the performance of a multi-layered network by carrying out a Layer sequential supervised initialization of  weights by exploring the high dimension error landscape using a stochastic method based on RMT, to be followed by a standard stochastic gradient descent (SGD) method for learning. We believe that our approach/solution is indeed based on the more general phenomena of finding the minima on a high dimension error landscape as conjectured by LeCun \cite{Lecun}. Authors in ~\cite{hessian} suggest that the method of second-derivatives intrinsically overcomes issues with pathological curvature during optimization. They also show that pre-trained initialization does not benefit the problem of under-fitting. However our method provides probabilistic guarantees on the nature of the minima the optimization converges to. This would lead us to believe that our process of weight initialization should improve the efficiency of any optimization algorithm that is used for training Neural Networks.

\section{Exploring high dimensional landscapes} \label{sec: theory}
In this section we model the error function of NN as a random landscape defined in equation \ref{eqn:ham1}. To find the minima of this high dimensional landscape, which are the optimal initial weights for gradient descent, we propose a method based on RMT. We compute weights using our RMT theory which are used to initialize the synaptic weights of NN prior to back propagation.  We consider an application of Face Recognition (FR) for analysis and justification of our approach. 

\subsection{Face Recognition} In FR, the neural network is trained to identify input images as one of the categories/classes in the training database. A detailed description of FR is given in \cite{FR}. Initially we consider a toy model of a single layer of neurons with $N$ input nodes and $N_{class}$ output nodes, where $N$ is the dimension of input sample and $N_{class}$ is the number of classes/categories in the training database.  For simplicity, initially we do not consider the non linear activation function, $f$, at the output nodes. Each link connecting $i^{th}$ input node to $j^{th}$ output node is associated with a synaptic weight $w_{i,j}$. The output of each output node is given by
\begin{equation} \label{eqn: output_node}
y_j = \sum_{i = 1}^{N}w_{i,j}x_{i}, \hspace{1cm} j = 1,2,...N_{class}
\end{equation}

During training of this network, samples from the training database are applied at the input nodes. The synaptic weights are tuned according to the difference between the network output $y$ and the desired output $y^d$.

From equation \ref{eqn: output_node}, error at the output node $j$ is given by,
\begin{equation}\label{eqn:err1}
E_{j} = \left(y^d_{j} - \sum_{i = 1}^{N}w_{i,j}x_{i}\right)^{2}
\end{equation} 

Suppose we write the desired output $y^d$ in the form
\begin{equation}\label{eqn:err2}
y^d_{j} = \sum_{i = 1}^{N}u_{i,j}
\end{equation}
where $u_{i,j}$ is a real number. Substituting this value of $y^d_{j}$ from equation \ref{eqn:err2} back into \ref{eqn:err1} we get
\begin{multline}
\label{eqn:err3}
 E_{j} = \left(\sum_{i = 1}^{N}u_{i,j}\right)^{2} - 2\left(\sum_{i = 1}^{N}u_{i,j}\right)\left(\sum_{i = 1}^{N}w_{i,j}x_{i}\right) \\ + \left(\sum_{i = 1}^{N}w_{i,j}x_{i}\right)^{2}  
\end{multline}

Expanding the first term in the R.H.S, we get
\begin{multline}
\label{eqn:err4}
 E_{j} = \sum_{i = 1}^{N}u_{i,j}^{2} + \sum_{i=1}^N \sum_{\substack {k=1 \\ i \neq k}}^N \left(u_{i,j}u_{k,j}\right) + \cdots \\ - 2\left(\sum_{i = 1}^{N}u_{i}\right)\left(\sum_{i = 1}^{N}w_{i,j}x_{i}\right)  + \left(\sum_{i = 1}^{N}w_{i,j}x_{i}\right)^{2}  
\end{multline}
The training of $N_{class}$ output nodes is independent of each other. As our target is to arrive at a favourable initial synaptic weights for gradient descent, we believe that these weights can be computed by training each output node with samples from the corresponding class. 

Consider the following interpretation. Let ${x^{a}}$ represent a sample that belongs to the training dataset of our toy network. Let it belong to the $a^{th}$ class of samples and therefore let the sum 
\begin{equation}
\sum_{i = 1}^{N}w_{i,j}x_{i}^{a} = \sum_{i = 1}^{N}u_{i,j}^{a}
\label{eqn:err5}
\end{equation}
which by definition is equal to $y_{j}^{a}$. Similarly for every sample $\bf{x^{k}}$ in the training dataset we can associate with it a vector $\bf{u^{k}}$ that maps it to the correct output node. 

The first term in the R.H.S of equation \ref{eqn:err4} is quadratic in $\bf{u}$. The other terms in the R.H.S are all complicated terms in $\bf{x}$ and $\bf{u}$. Observe that we can always define a transformation between $x_{i}$ and $u_{i,j}$ in $N$-dimensions. This is true because both coordinates represent different ways of looking at the same landscape. $x_{i}$s are defined over a \lq position-field\rq  , whereas $\bf{u}$s are defined over the \lq transform-field\rq. Thus the error function is the sum of a quadratic term in $\bf{u}$ and a complicated $N$-dimensional function of variables $\bf{u}$.

Since we are already in $N$-dimensions and are dealing with an error function that has complicated terms, consider applying the following trick. We replace the complicated part of the error function,$E_j$ by a random function $V(u_{1,j}, u_{2,j}, \cdots, u_{N,j})$ in $N$-dimensions where $u_{i,j}$ are mean-zero Gaussian random variables. This assumption allows us to apply the tools of Random Matrix Theory to analyze the error function as we will discuss in section \ref{sec: theory1}

The above assumption begs the following question: How do we know that are coordinates(images) are indeed mean-zero Gaussian variables? We look to statistics for an answer to this query. Each $u_{j}$ represents a pixel of a face image in the dataset. Given a large database, by the Central Limit Theorem (C.L.T.) each pixel takes Gaussian values. This answers the Gaussian part, now let's consider the mean-zero part. Each pixel is associated with Gaussian values and a mean. Consider the mean of all the pixels of an image. High resolution images have lots of pixels. Therefore the mean of each pixel must also conform to a Gaussian by the C.L.T. If we redefine the origin of coordinates to this number, then provided that the means have low variance, we have succeeded in producing $N$ independent mean-zero Gaussian random variables. Observe that we have introduced a constraint by way of requiring that the means have a low variance. We can ensure this constraint by considering only similar images, i.e., images belonging to same class, because similar images by definition don't \lq vary\rq{\ }much from each other. We would like to impress that such a constraint doesn't effect the generality of our results. This is because we choose to train each output node of the neural network independently. 

\subsection{Finding Minima on a High Dimensional error Landscape} \label{sec: theory1}

To find the minima of the error function in equation \ref{eqn:err4}, we look for analogy between the error function and N dimensional G.O.E. in equation \ref{eqn:ham1}. By representing terms other than the first term in RHS of equation \ref{eqn:ham1} as a random function $V(x)$, we write it as an $N$-dimensional Hamiltonian of the form

\begin{equation}
\label{eqn:ham11}
\mathcal{H} = \frac{\mu}{2}\sum_{i = 1}^{N}u_{i}^{2} + V(u_{1}, u_{2}, ..., u_{N})
\end{equation} 
where $\mu$ is a tuning parameter and $V$ is a $N$-dimensional random landscape with covariance given by 
equation \ref{eqn:ham2}. We see that $V(u_{1}, u_{2}, ..., u_{N})$ is indeed a very complex function which can be approximated to be a mean-zero Gaussian-distributed field. We assume that the $\bf{u}$ follows the same distribution as that of $\bf{x}$ that helps us in finding the minima of the error landscape. As the training samples are face images which are rotation and translation invariant, we can assume the covariance matrix also to be isotropic. Having confirmed the basic requirements, we can apply the RMT method described in section \ref{sec: RMT}.

In our approach, we compute weight vectors connecting each output node independently by using samples of corresponding class. Thus, we compute $N_{class}$ N-dimensional landscape minima to train our toy network. From equation \ref{eqn:fyod5}, \ref{eqn:dens1} and \ref{eqn:dens2} we conclude that the mean number of energy minima of $\mathcal{H}$ is a function of $h_N$. 

\begin{figure}[!b]
  \centering
 \includegraphics[height=5.0cm, width=7cm]{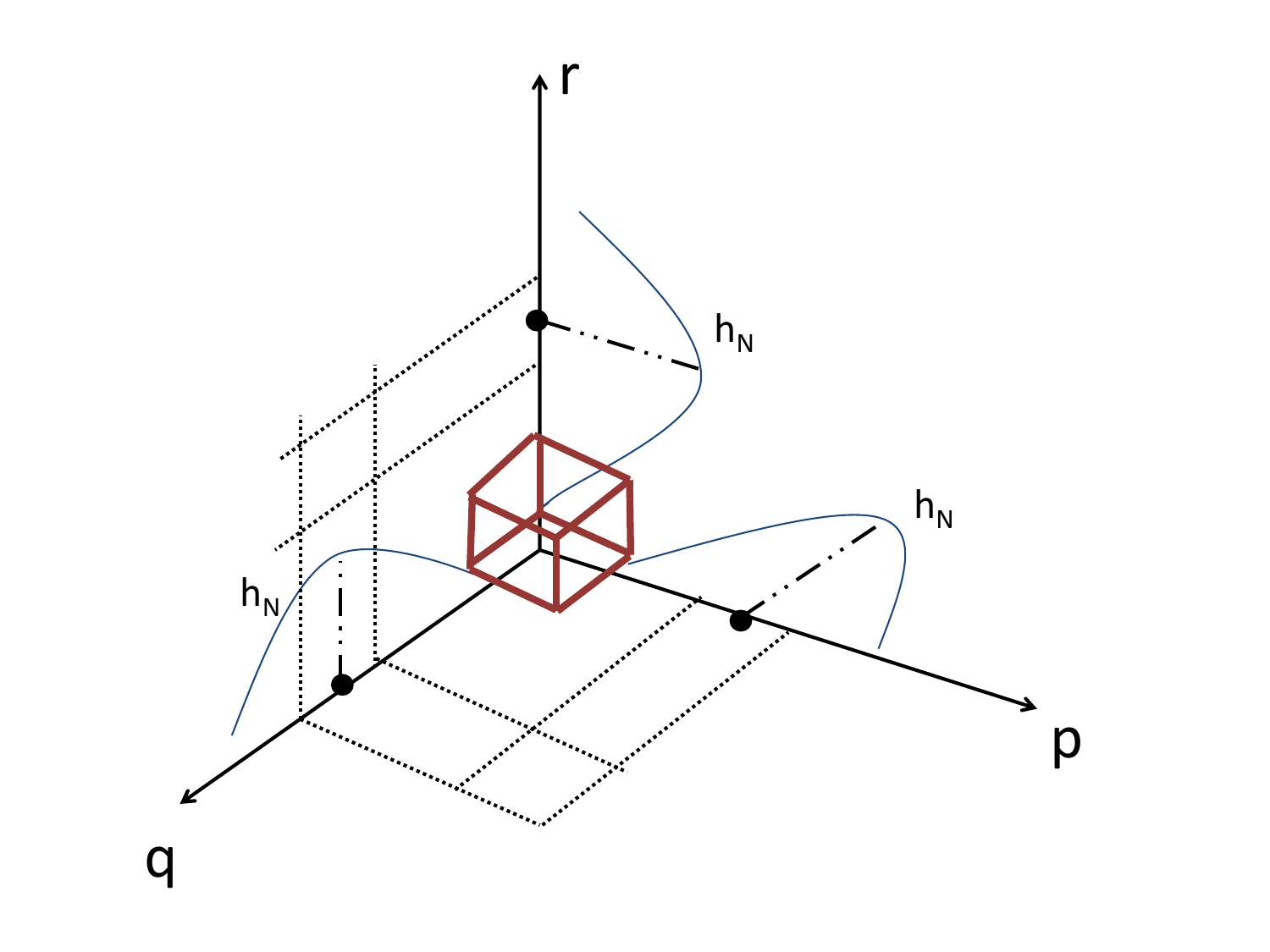} 
\caption{Construction of a cube based on maximum point of $h_N$ in a three dimensional case.}
  \label{fig: cube}
  \end{figure}

To train an output node, the required solution is the value of $\bf{u}$ that will have maximum mean number of minima indicating the presence of optimum point. Looking at equation \ref{eqn:fyod5} we find that $h_N$ is a function of $\frac{\mu}{\mu_c}$, where ${\mu_c}$ is equal to $\sqrt{f^{\prime\prime}(0)}$, and f(x) is the covariance function. The last term in equation \ref{eqn:fyod5} is logarithm of derivative of solution of Painlev\'{e} II equation that is observed to be negligible when compared to the difference of first two terms for large $N$. Therefore, in equation \ref{eqn:fyod5}, $\mu_c$ behaves as a critical point in every dimension representing the point where we can find majority of minima. Gradient is computed on the covariance matrix two times and square root of the diagonal elements are assigned to $\mu$s in respective dimensions. In equations \ref{eqn:fyod5}, \ref{eqn:dens1} and \ref{eqn:dens2}, it is observed that to find average number of minima in a spatial domain, $h_{N}$ is integrated over the range in each dimension. A randomly selected vector from the hypercube that shows a high density of minima will give us the solution vector $\bf{u}$. This implies that to find our initialization weights, it would suffice to find in each dimension $\bf{u}$ for which $h_N$ becomes maximum. Fig. \ref{fig: cube} shows a three dimensional case where a cube is constructed based on the max of $h_N$ in each dimension. Finally the computed $\bf{u}$ is scaled down by the range of input pixels to get the synaptic weights.

In the section \ref{sec: analysis}, we analyse our method from an implementation perspective and select the parameters to achieve accurate and robust classification. In addition, we extend this work to multi-layer weight initialization with promising results.

\section {Analysis and results} \label{sec: analysis}

For functional verification of our RMT based weight initialization method, we use two face image databases: AR and Extended Yale database B. AR database consists of 100 classes each containing 26 face images. Extended Yale database B consists of 37 classes, each with 20 face images. The images are preprocessed, as described in section \ref{sec: prepro}, before using them for weight computation. We use Matlab NN tool \cite{matlab} to verify the effectiveness of our weights. By default, Matlab NN tool uses Nguyen-Widrow weight initialization method \cite{NW} and Conjugate Gradient Descent \cite{conjugate} for training the weights. We apply the weights computed using our RMT based method to the network before training and analyse and compare the performance in terms of number of epochs for convergence and final recognition accuracy of NN. We see different recognition accuracies and number of epochs when the NN tool is run multiple times with the same initialization weights. Therefore, we run the same experiment multiple times and report the arithmetic mean, and maximum values in this paper.

\subsection{Pre-processing the training data}\label{sec: prepro}
We pre-process the samples by reducing the dimension and also mean centring them. We define a G.O.E. of input obtained by performing Principle Component Analysis (PCA) \cite{PCA} on the test samples of faces. The eigenspace defined for the faces not only serves to reduce the dimension of the inputs, but also helps defining an error landscape over a G.O.E. of principal components/axes.  It is also possible to carry out Feature Extraction (PCA) following the Random Matrix Theory approach as reported in \cite{feature_random} for dimension reduction. Resultant feature vectors are mean centred and divided by their standard deviation to make it mean-zero and unit variance.


\subsection{Selection of $\mu$}
\begin{figure}[!b]
  \centering
 \includegraphics[height=4.0cm, width=6cm]{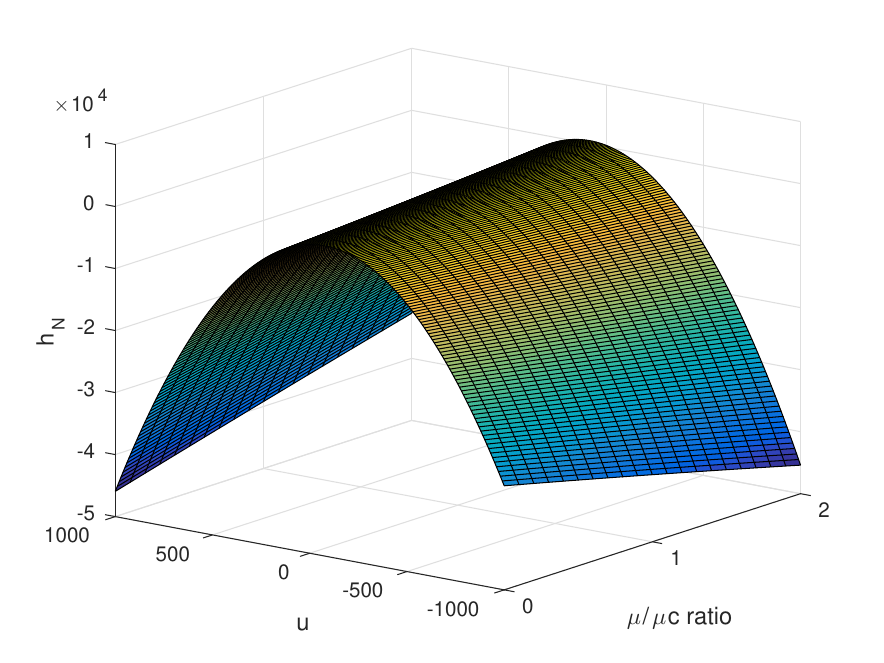} 
\caption{A 3D plot showing variation of $h_{N}$ with $\frac{\mu}{\mu_c}$ ratio and $\bf{u}$ for $N$ equal to 256}
  \label{fig: 3d}
  \end{figure}

\begin{figure}[!b]
  \centering
 \includegraphics[height=4.0cm, width=6cm]{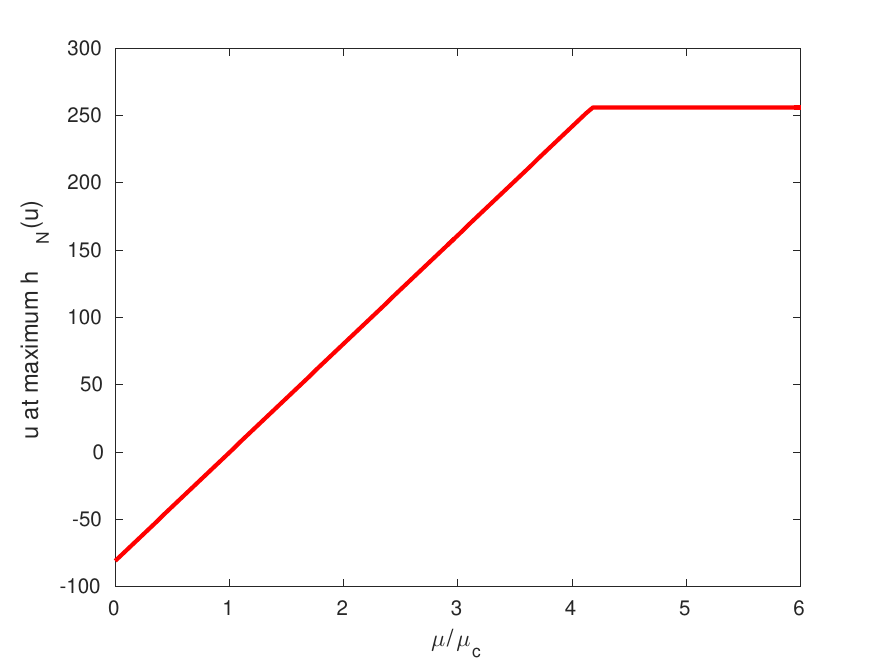} 
\caption{saturation of $\bf{u}$ for high $\frac{\mu}{\mu_c}$ }
  \label{fig: mmc}
  \end{figure}

From section \ref{sec: theory1} it is clear that $\mu_c$ is the critical parameter which is responsible for variations in $h_N$ for a fixed $N$. A three dimensional plot in Fig. \ref{fig: 3d} shows variation of $h_N$ with different values of $\bf{u}$ and $\frac{\mu}{\mu_c}$. We get components of $\bf{u}$ in different dimensions where $h_N$ becomes maximum based on the $\frac{\mu}{\mu_c}$ ratio. Fig. \ref{fig: mmc} shows a plot of indices of $\bf{u}$ for maximum $h_N$ with varying $\frac{\mu}{\mu_c}$. We find that the plot saturates beyond a point after which we will have the same computed values of $\bf{u}$.
Our aim is to get the values of $u$ in the linear region of the plot so that they cover the whole query region of the landscape for effective weight computation. Therefore we select the tuning parameter $\mu$ such that the ratio $\frac{\mu}{\mu_c}$ falls in the required range. 


%

\subsection{Multi-layer network weight computation}
Until now we have established our theory for our toy model of network which is a single layer of neurons. When it comes to training a network for a large face database with complex features, single layer of neurons are not sufficient to achieve the required classification performance. To make our theory applicable to these conditions, we extend our theory to multi-layer networks which requires computation of more than one set of weights. In multi-layer neural networks, synaptic weights associated with neurons in hidden layers try to identify patterns in the input samples. These patterns will be used collectively for classification in subsequent layers. During back propagation, the weights tune themselves for identifying a pattern in the input samples. This behaviour of hidden nodes help us in extending our theory to multi-layer networks. Here, similar to our toy model, we do not use bias inputs in all the layers.

\begin{figure}[t]
  \centering
 \includegraphics[height=6.0cm, width=7cm]{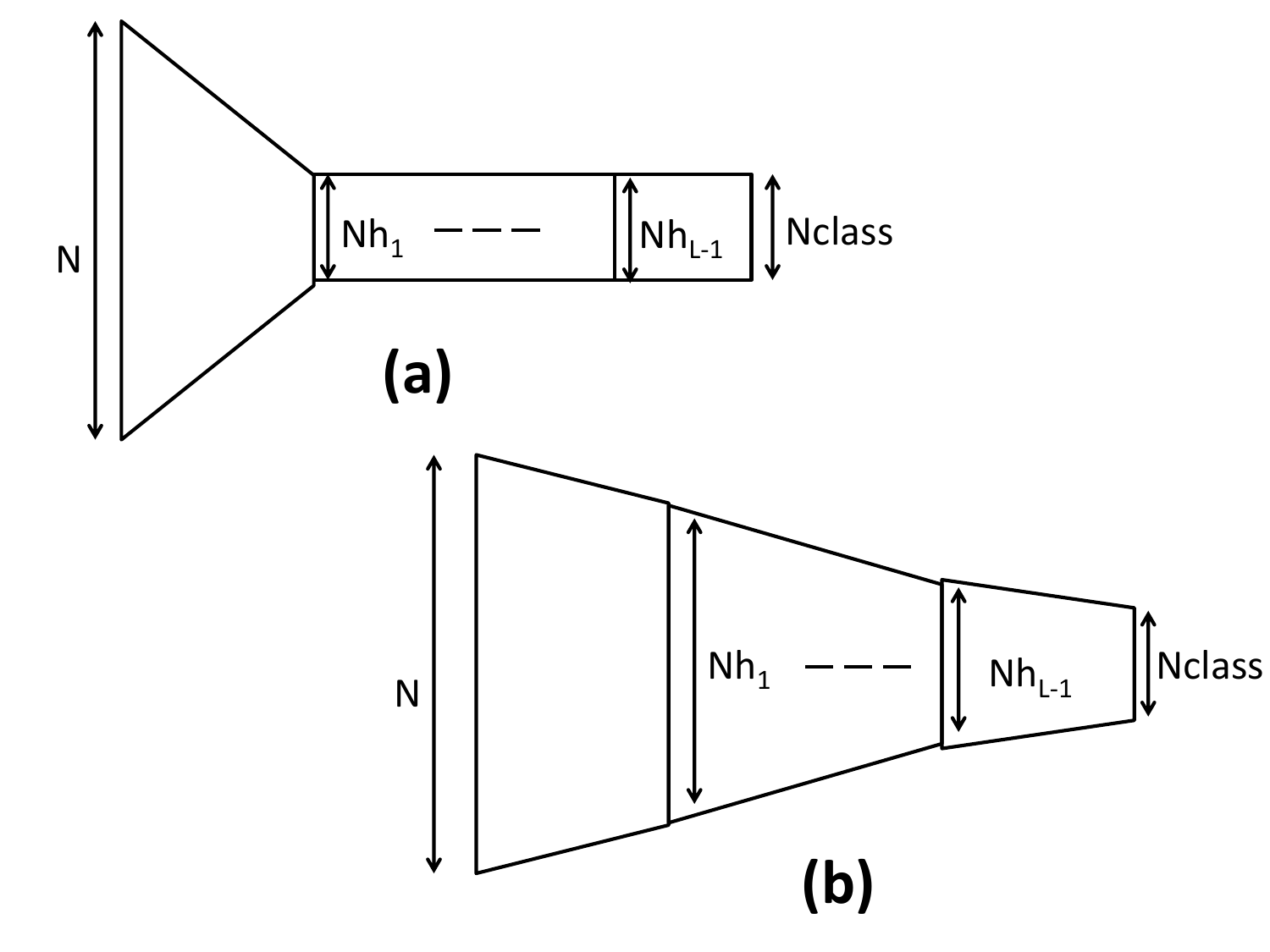} 
\caption{Multi-layer networks with (a) equal and (b)unequal number of neurons in $L$ number of layers}
  \label{fig: multi}
  \end{figure}

\begin{figure}
\subfigure[Recognition accuracy for AR face database\label{fig: AR_rec}]{\includegraphics[width=7cm, height=4.0cm]{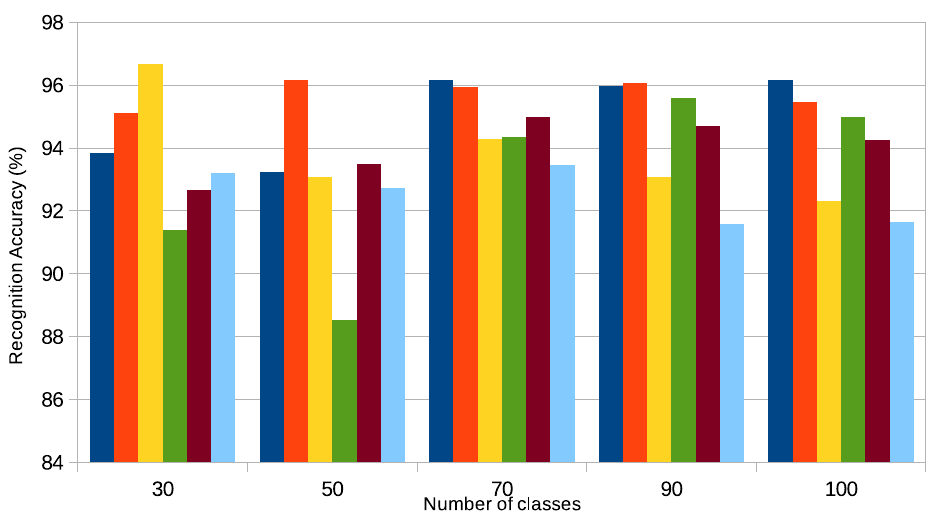}}\\
\subfigure[Recognition accuracy for Extended Yale face database B\label{fig: yale_rec}]{\includegraphics[width=7cm, height=4.0cm]{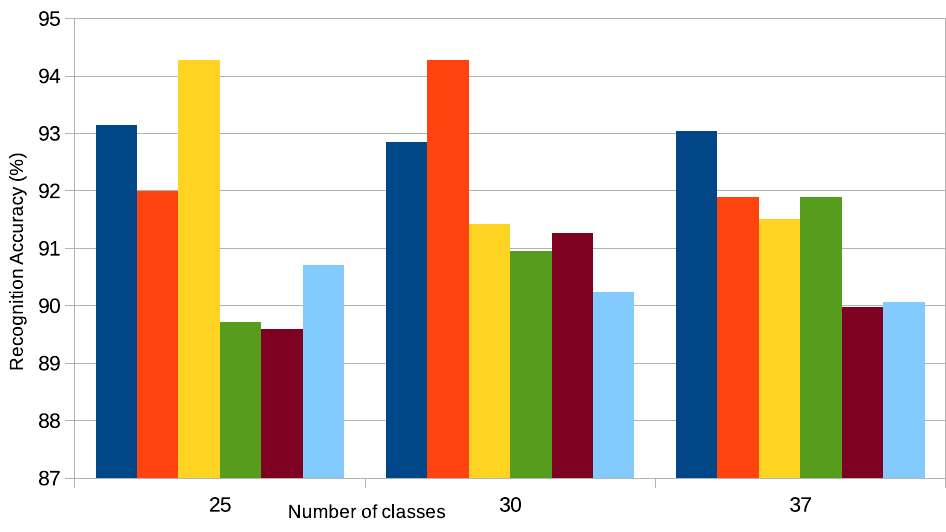}} 
\subfigure[Number of epochs for convergence for AR face database  \label{fig: AR_iter}]{\includegraphics[width=7cm, height=4.0cm]{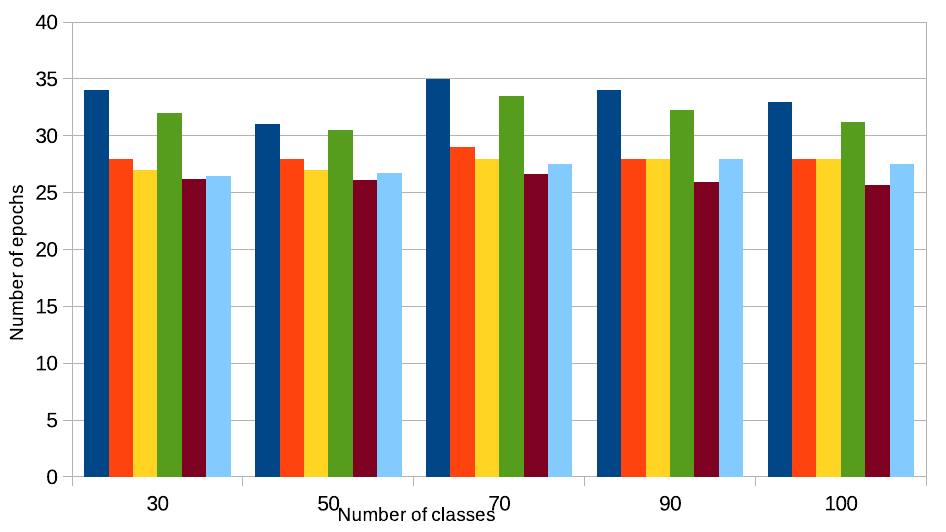}}\\
\subfigure[Number of epochs for convergence for Extended Yale face database B
 \label{fig: yale_iter}]{\includegraphics[width=7cm, height=4.0cm]{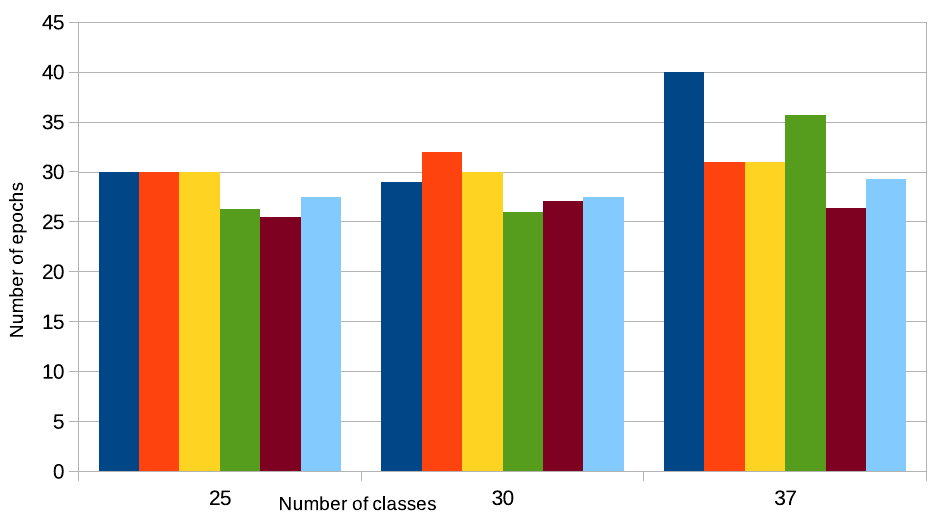}}\\
\subfigure{\includegraphics[width=5cm, height=0.8cm]{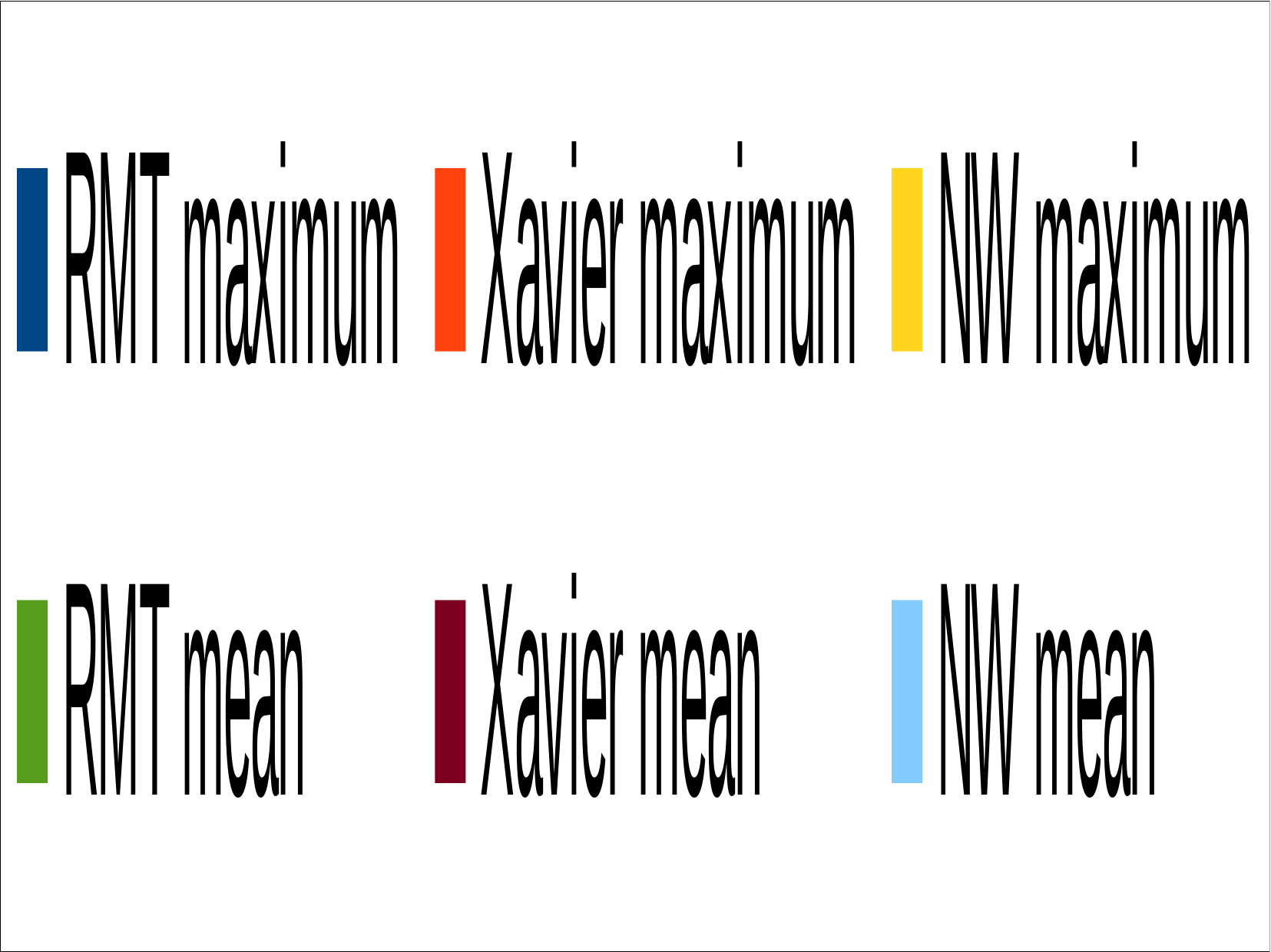}}\\
 
\caption{Performance of RMT based weight initialization method for two-layer network with equal number of nodes in layers}
\label{fig: plots1}
\end{figure}

In multi-layer networks, application of our method is not straight forward. Our method relies on class-wise computation of weights, that needs equal number of nodes in each hidden layer and the output layer as shown in Fig. \ref{fig: multi}(a). Let $N$ and $Nh_{i}$ are the number of nodes in the input and $i^{th}$ hidden node respectively. Initially we apply our method on first layer of neurons and compute a weight matrix of dimension $N \times Nh_1$. This is analogous to training each first layer hidden node with samples from respective class. This may look like an unorthodox way of training, but we believe that this will help to achieve better weight initialization. Once the weight matrix for the first layer is computed, we compute output of the first hidden layer by multiplying with the input sample set $N_{samp} \times N$ and a sigmoid operation. Although our theory does not include a sigmoid unit at the neuron outputs, we have experimentally observed  that the computed weights performed well with sigmoid units included in the network. The computed outputs of dimension $N_{samp} \times Nh_1$ from first hidden layer act as inputs for the second layer with associated labels. Similarly we compute synaptic weights for rest of the hidden layers and the output layer. Previously in our toy model of NN, synaptic weights of a neuron were computed considering only samples of respective class. However, our method when extended to a multi-layer NN, brings this indirect dependency which will improve the classification performance. e.g., first node in the second layer gets inputs from all the nodes in the previous layer and thus, is dependent on the samples used under  those nodes in computing the weights. Plots in Fig. \ref{fig: plots1} compare recognition accuracy and number of epochs for convergence for our weight initialization with NW initialization method used by Matlab NN tool. We use the maximum and average values over a number of runs for this comparison. In Fig. \ref{fig: AR_rec} and Fig. \ref{fig: yale_rec} we observe that for large number of classes, on an average, our method of weight initialization has resulted in better recognition accuracy. In addition, we see that the difference between recognition accuracy of our method and that of other standard methods increases with number of classes. Finding better minima in the error landscape may result in additional steps of weight updates which is observed in Fig. \ref{fig: AR_iter} and Fig. \ref{fig: yale_iter}. 

\begin{figure}
\subfigure[Recognition accuracy for 100 classes of AR face database\label{fig: AR_rec1}]{\includegraphics[width=7cm, height=4.0cm]{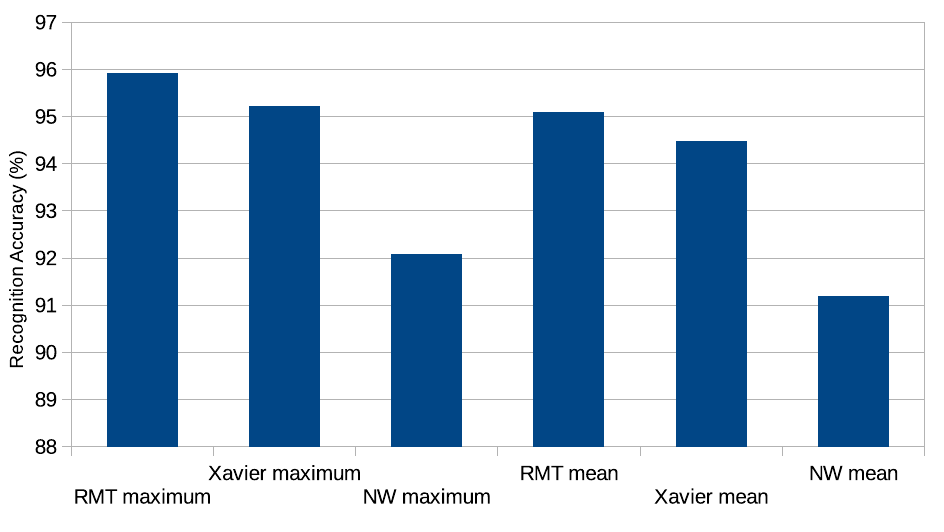}}\\
\subfigure[Number of epochs for convergence for 100 classes of AR face database  \label{fig: AR_iter1}]{\includegraphics[width=7cm, height=4.0cm]{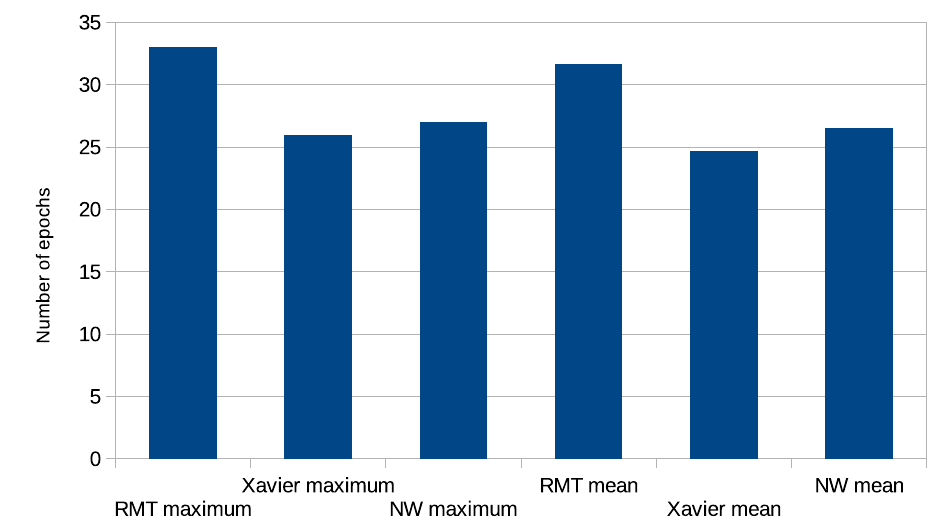}}\\
\caption{Performance of RMT based weight initialization method for two-layer network with unequal number of nodes in layers (150 nodes in the hidden layer)}
\label{fig: plots2}
\end{figure}

The multi-layer network described above has a limitation that the hidden and the output layer must have $N_{class}$ number of nodes. In practical networks this is not a valid scenario, as the number of nodes in each layer differs based on the training database and available resources. In addition, for inputs with very large dimensions and small number of classes, the network cannot represent the classification function efficiently. We address this issue by using our RMT method with a clustering algorithm introduced at each layer. We use k-means clustering which is an unsupervised clustering algorithm for our experiments. The samples are clustered by the algorithm based on their similarity and the target number of clusters. To apply our method on this set-up, we consider each cluster as a single class. We believe that it is a valid assumption as the samples in a cluster are similar to each other due to adjacency in the feature space. In addition, the hidden similar patterns in samples from different classes are explicitly brought together under a neuron. Fig. \ref{fig: multi}(b) shows a NN with varying number of nodes in each layer. Performance of multi-cluster network on 100 classes of AR database with different methods of weight initializations is shown using plots in Fig. \ref{fig: AR_rec1} and Fig. \ref{fig: AR_iter1}. Here too we observe that our RMT based initialization method outperforms other methods in terms of recognition accuracy.

\subsection{Significance of RMT based weight initialization in real-life learning problems}

The present work focuses on finding initial weights for multi-layered shallow networks. The same theory applies to deeper networks and convolution networks which we will target in our future work. In shallow networks, we have analyzed performance of our algorithm with respect to number of epochs and recognition accuracy for face recognition application. Although we have shown the performance on this single application with few standard face databases, we believe that the algorithm behaves similarly for other recognition applications where the database obeys the conditions necessary for applicability of RMT mentioned in section \ref{sec: theory}. All natural image recognition problems fall under this category. 

We have observed that an NN with our RMT based weight initialization method results in better recognition accuracy as compared to that of NN with standard weight initializations. Although relatively standard initializations are lower in complexity, the investment in additional computations brings in returns by way of higher recognition accuracy. The computations involved in finding initial weights per layer using RMT are (not considering PCA which is a pre-processing step) covariance computation($O(N^2N_{samp})$) and search for maximum value ($O(Nlog N)$). When the number of nodes in each layer are not equal to the number of classes, the additional computations involve k-means clustering($O(N_{samp}^{kN+1}log N_{samp})$, $k$=number of clusters). We intend to replace the k-means clustering with a relatively simpler clustering method in our future work. 

In the case of large databases, typically training is performed off-line. Such learning systems are hosted on multi-core platforms due to latent parallelism in the training algorithm. The additional computations required for RMT based weight initialization also can be sufficiently parallelized on these platforms and hence they do not significantly impact the overall training time. In Internet of Things (IoT) devices, for on-line training, size of the database is much smaller. For such cases, better recognition accuracy is achieved by our method as a trade-off for cost of additional computations.



\subsection{Future direction in rigorous mathematical validation}
\begin{figure}[!b]
  \centering
 \includegraphics[height=6cm, width=8cm]{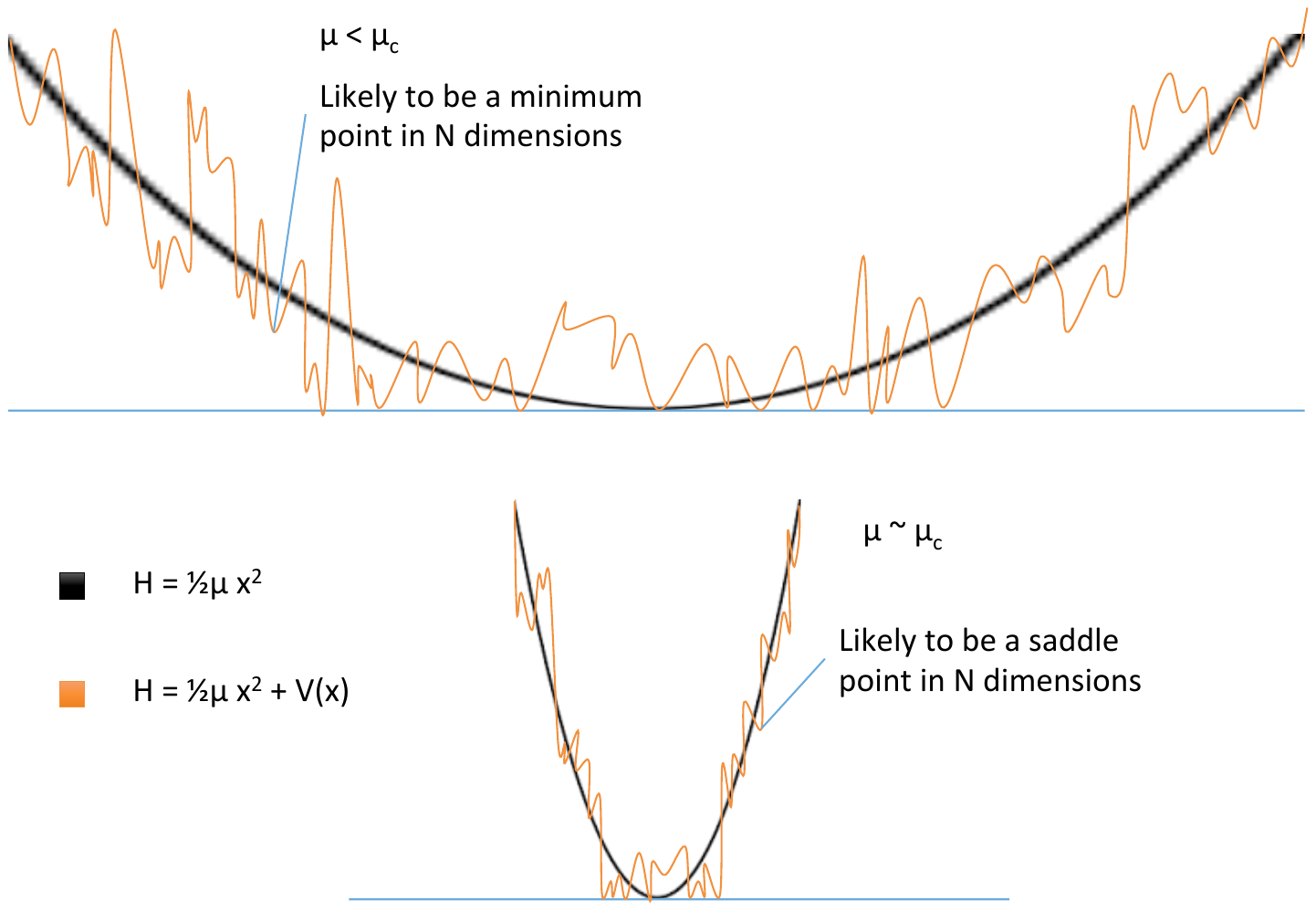} 
\caption{Nature and density of minima at different values for $\mu$. In the case that $\mu < \mu_c$, the landscape is characterized by exponential number of minima. This is because the parabola defined by a small value for $\mu$ tends to be broad. The randomness in the potential V(x) therefore tends to produce a lot of bad minima. In the case that $\mu \sim \mu_c$, the parabola is sharper and the number of minima is sub-exponential.}
  \label{fig:mupot}
  \end{figure}

In our quest to get the best possible initialization, we choose $\mu$s so as to get empirically optimal values for $\mu/\mu_c$.  This presents an uneasy situation that suffers from the lack of theoretical validation. However, the following schematic provides (see Fig. \ref{fig:mupot}) some insight as to why our choices for the $\mu$s works.

Existing literature \cite{fyod1} talks about a phase transition in the energy landscape of the Hamiltonian (as defined in equation \ref{eqn:ham1}). The phase when $\mu < \mu_c$ has exponential number of minima and the phase when $\mu > \mu_c$ has just one minima. The phase transition region (at a distance $\delta$ about $\mu_c$) is characterised by a sub-exponential number of minima. We conjecture that this phase transition region is defined by our choice of $\mu$. Therefore our search for minima on this landscape is likely to find true minima. This is opposed to the case where there are exponential number of minima and gradient decent leads us to converge into bad minima. We believe that this paper presents a new way of developing robust neural networks. The authors plan to validate this conjecture in the immediate future. 


\section {Conclusion}  \label{sec: conclusion}

Accurate classification is the goal of any multi-layer large neural network. Initialization of weights has a significant impact on the convergence of learning algorithms. We have provided a statistical method based on Random Matrix Theory for the weight initialization with probabilistic guarantees on the nature of minima reached in the high dimension landscape defined by the error function. Experimentally we have substantiated the novelty of our method in obtaining higher classification accuracy over well known initialization methods adopted by deep learning frameworks.

\section*{Acknowledgment}
The authors would like to thank Chandrasekhar Seelamantula, S.K. Nandy and Ranjani Narayan for their valuable suggestions in this work

\bibliographystyle{IEEEtran}
\bibliography{IEEEabrv,references}

\begin{thebibliography}{10}
\providecommand{\url}[1]{#1}
\csname url@samestyle\endcsname
\providecommand{\newblock}{\relax}
\providecommand{\bibinfo}[2]{#2}
\providecommand{\BIBentrySTDinterwordspacing}{\spaceskip=0pt\relax}
\providecommand{\BIBentryALTinterwordstretchfactor}{4}
\providecommand{\BIBentryALTinterwordspacing}{\spaceskip=\fontdimen2\font plus
\BIBentryALTinterwordstretchfactor\fontdimen3\font minus
  \fontdimen4\font\relax}
\providecommand{\BIBforeignlanguage}[2]{{%
\expandafter\ifx\csname l@#1\endcsname\relax
\typeout{** WARNING: IEEEtran.bst: No hyphenation pattern has been}%
\typeout{** loaded for the language `#1'. Using the pattern for}%
\typeout{** the default language instead.}%
\else
\language=\csname l@#1\endcsname
\fi
#2}}
\providecommand{\BIBdecl}{\relax}
\BIBdecl

\bibitem{init}
\BIBentryALTinterwordspacing
D.~Mishkin and J.~Matas, ``All you need is a good init,'' \emph{CoRR}, vol.
  abs/1511.06422, 2015. [Online]. Available:
  \url{http://arxiv.org/abs/1511.06422}
\BIBentrySTDinterwordspacing

\bibitem{xavier}
X.~Glorot and Y.~Bengio, ``Understanding the difficulty of training deep
  feedforward neural networks,'' in \emph{In Proceedings of the International
  Conference on Artificial Intelligence and Statistics (AISTATS’10). Society
  for Artificial Intelligence and Statistics}, 2010.

\bibitem{NW}
D.~Nguyen and B.~Widrow, ``Improving the learning speed of 2-layer neural
  networks by choosing,'' in \emph{Initial Values of the Adaptive Weights,
  International Joint Conference of Neural Networks}, 1990, pp. 21--26.

\bibitem{caffe}
\BIBentryALTinterwordspacing
Y.~Jia, E.~Shelhamer, J.~Donahue, S.~Karayev, J.~Long, R.~Girshick,
  S.~Guadarrama, and T.~Darrell, ``Caffe: Convolutional architecture for fast
  feature embedding,'' in \emph{Proceedings of the 22Nd ACM International
  Conference on Multimedia}, ser. MM '14.\hskip 1em plus 0.5em minus
  0.4em\relax New York, NY, USA: ACM, 2014, pp. 675--678. [Online]. Available:
  \url{http://doi.acm.org/10.1145/2647868.2654889}
\BIBentrySTDinterwordspacing

\bibitem{matlab}
``Neural network toolbox,''
  \url{http://in.mathworks.com/products/neural-network/}, accessed: 2016-03-07.

\bibitem{deep}
\BIBentryALTinterwordspacing
G.~E. Hinton, S.~Osindero, and Y.-W. Teh, ``A fast learning algorithm for deep
  belief nets,'' \emph{Neural Comput.}, vol.~18, no.~7, pp. 1527--1554, jul
  2006. [Online]. Available:
  \url{http://dx.doi.org/10.1162/neco.2006.18.7.1527}
\BIBentrySTDinterwordspacing

\bibitem{vanishing}
\BIBentryALTinterwordspacing
S.~Hochreiter, ``The vanishing gradient problem during learning recurrent
  neural nets and problem solutions,'' \emph{Int. J. Uncertain. Fuzziness
  Knowl.-Based Syst.}, vol.~6, no.~2, pp. 107--116, Apr. 1998. [Online].
  Available: \url{http://dx.doi.org/10.1142/S0218488598000094}
\BIBentrySTDinterwordspacing

\bibitem{Edelman}
A.~Edelman and Y.~Wang, \emph{Advances in Applied Mathematics, Modeling, and
  Computational Science}.\hskip 1em plus 0.5em minus 0.4em\relax Boston, MA:
  Springer US, 2013, ch. Random Matrix Theory and Its Innovative Applications,
  pp. 91--116.

\bibitem{wigner}
\BIBentryALTinterwordspacing
O.~R. N. L.~P. Division and U.~A.~E. Commission, \emph{Conference on Neutron
  Physics by time-of-flight, held at Gatlinburg, Tennessee, November 1 and 2,
  1956}.\hskip 1em plus 0.5em minus 0.4em\relax Oak Ridge National Laboratory,
  1957. [Online]. Available:
  \url{https://books.google.co.in/books?id=k0hwfgVnEEkC}
\BIBentrySTDinterwordspacing

\bibitem{majumdar}
S.~N. Majumdar, \emph{https://www.icts.res.in/media/uploads/Program/
  Files/satya1.pdf}.\hskip 1em plus 0.5em minus 0.4em\relax Lecture notes,ICTS,
  2012.

\bibitem{tracy}
\BIBentryALTinterwordspacing
C.~A. Tracy and H.~Widom, ``Level-spacing distributions and the airy kernel,''
  \emph{Comm. Math. Phys.}, vol. 159, no.~1, pp. 151--174, 1994. [Online].
  Available: \url{http://projecteuclid.org/euclid.cmp/1104254495}
\BIBentrySTDinterwordspacing

\bibitem{fyod1}
\BIBentryALTinterwordspacing
Y.~V. Fyodorov and C.~Nadal, ``Critical behavior of the number of minima of a
  random landscape at the glass transition point and the tracy-widom
  distribution,'' \emph{Phys. Rev. Lett.}, vol. 109, p. 167203, Oct 2012.
  [Online]. Available:
  \url{http://link.aps.org/doi/10.1103/PhysRevLett.109.167203}
\BIBentrySTDinterwordspacing

\bibitem{mehta}
M.~mehta, \emph{random matrices}.\hskip 1em plus 0.5em minus 0.4em\relax
  Academic Press, 2004.

\bibitem{fitnet}
\BIBentryALTinterwordspacing
A.~Romero, N.~Ballas, S.~E. Kahou, A.~Chassang, C.~Gatta, and Y.~Bengio,
  ``Fitnets: Hints for thin deep nets,'' \emph{CoRR}, vol. abs/1412.6550, 2014.
  [Online]. Available: \url{http://arxiv.org/abs/1412.6550}
\BIBentrySTDinterwordspacing

\bibitem{deep_net}
\BIBentryALTinterwordspacing
R.~K. Srivastava, K.~Greff, and J.~Schmidhuber, ``Training very deep
  networks,'' in \emph{Advances in Neural Information Processing Systems 28},
  C.~Cortes, N.~D. Lawrence, D.~D. Lee, M.~Sugiyama, and R.~Garnett, Eds.\hskip
  1em plus 0.5em minus 0.4em\relax Curran Associates, Inc., 2015, pp.
  2368--2376. [Online]. Available:
  \url{http://papers.nips.cc/paper/5850-training-very-deep-networks.pdf}
\BIBentrySTDinterwordspacing

\bibitem{greedy}
\BIBentryALTinterwordspacing
Y.~Bengio, P.~Lamblin, D.~Popovici, and H.~Larochelle, ``Greedy layer-wise
  training of deep networks,'' in \emph{Advances in Neural Information
  Processing Systems 19}, B.~Sch\"{o}lkopf, J.~C. Platt, and T.~Hoffman,
  Eds.\hskip 1em plus 0.5em minus 0.4em\relax MIT Press, 2007, pp. 153--160.
  [Online]. Available:
  \url{http://papers.nips.cc/paper/3048-greedy-layer-wise-training-of-deep-networks.pdf}
\BIBentrySTDinterwordspacing

\bibitem{Lecun}
``Explorations on high dimensional landscapes,'' \url{http:
  //arxiv.org/abs/1412.6615?context=stat.ML}, accessed: 2016-03-07.

\bibitem{hessian}
J.~Martens, ``Deep learning via hessian-free optimization.''

\bibitem{FR}
G.~Mahale, H.~Mahale, S.~Nandy, and N.~Ranjani, ``Refresh: Redefine for face
  recognition using sure homogeneous cores,'' \emph{IEEE Transaction on
  Parallel and Distributed Systems}, 2016.

\bibitem{conjugate}
``A bried introduction to the conjugate gradient method,''
  \url{http://www.idi.ntnu.no/~elster/tdt24/tdt24-f09/cg.pdf}, accessed:
  2016-03-07.

\bibitem{PCA}
\BIBentryALTinterwordspacing
M.~Turk and A.~Pentland, ``Eigenfaces for recognition,'' \emph{J. Cognitive
  Neuroscience}, vol.~3, no.~1, pp. 71--86, jan 1991. [Online]. Available:
  \url{http://dx.doi.org/10.1162/jocn.1991.3.1.71}
\BIBentrySTDinterwordspacing

\bibitem{feature_random}
V.~Rojkova and M.~Kantardzic, ``Feature extraction using random matrix theory
  approach,'' in \emph{Machine Learning and Applications, 2007. ICMLA 2007.
  Sixth International Conference on}, Dec 2007, pp. 410--416.

\end{thebibliography}



%
%
%

%
%

\end{document}